\documentclass[runningheads]{llncs}

\usepackage{xcolor}
\usepackage{booktabs}
\usepackage{url}
\usepackage{lmodern}
\usepackage{amsmath, amssymb}
\usepackage{graphicx}

\begin{document}

\title{Meta-learning and Data Augmentation for Stress Testing Forecasting Models\thanks{This work was partially funded by projects AISym4Med (101095387) supported by Horizon Europe Cluster 1: Health, ConnectedHealth (n.º 46858), supported by Competitiveness and Internationalisation Operational Programme (POCI) and Lisbon Regional Operational Programme (LISBOA 2020), under the PORTUGAL 2020 Partnership Agreement, through the European Regional Development Fund (ERDF) and NextGenAI - Center for Responsible AI (2022-C05i0102-02), supported by IAPMEI, and also by FCT plurianual funding for 2020-2023 of LIACC (UIDB/00027/2020 UIDP/00027/2020)}}

\titlerunning{Meta-learning and Data Augmentation for Stress Testing Forecasting Models}

\author{Ricardo~Inácio\inst{1,2}\orcidID{0009-0008-6435-5245}, Vitor~Cerqueira\inst{1,2}\orcidID{0000-0002-9694-8423}, Marília~Barandas\inst{3}\orcidID{0000-0002-9445-4809} \and Carlos~Soares\inst{1,2,3}\orcidID{0000-0003-4549-8917}}

\authorrunning{R. Inácio et al.}

\institute{Faculdade de Engenharia da Universidade do Porto, Porto, Portugal \\
\email{ricardoinacio@mailbox.org} \\
\email{\{csoares,vcerqueira\}@fe.up.pt} \and
Laboratory for Artificial Intelligence and Computer Science (LIACC), Portugal \and
Fraunhofer Portugal AICOS, Portugal\\
\email{marilia.barandas@aicos.fraunhofer.pt}
}

\maketitle 

\begin{abstract}

The effectiveness of univariate forecasting models is often hampered by conditions that cause them stress. A model is considered to be under stress if it shows a negative behaviour, such as higher-than-usual errors or increased uncertainty.
Understanding the factors that cause stress to forecasting models is important to improve their reliability, transparency, and utility.
This paper addresses this problem by contributing with a novel framework called \texttt{MAST} (Meta-learning and data Augmentation for Stress Testing). The proposed approach aims to model and characterize stress in univariate time series forecasting models, focusing on conditions where they exhibit large errors. In particular, 
\texttt{MAST} is a meta-learning approach that predicts the probability that a given model will perform poorly on a given time series based on a set of statistical time series features.
\texttt{MAST} also encompasses a novel data augmentation technique based on oversampling to improve the metadata concerning stress.
We conducted experiments using three benchmark datasets that contain a total of 49.794 time series to validate the performance of \texttt{MAST}. The results suggest that the proposed approach is able to identify conditions that lead to large errors. 
The method and experiments are publicly available in a repository.

\end{abstract}

\section{Introduction}

Time series forecasting is a relevant problem in various application domains, such as industry, healthcare, or finance. Accurate forecasts help reduce uncertainty about the future and foster a data-driven decision-making within organizations \cite{jain2017answers}.
However, the effective application of forecasting models is sometimes hampered by \textit{stress} conditions that reduce their performance and reliability. 

In machine learning, stress denotes conditions where a model exhibits a negative behaviour, such as higher-than-usual errors or increased uncertainty. 
Stress factors result in unreliable predictions, causing end-users to distrust models.
In the context of forecasting, stress can stem from factors such as i) data difficulty problems, for instance out-of-distribution samples or missing data, or ii) an inadequate inductive bias where the model fails to capture relevant patterns in specific time series. Stress testing aims to evaluate the reliability of machine learning approaches by identifying, modelling, or simulating challenging scenarios. This process provides actionable insights for improving models and decision-making by end-users. 

\begin{figure}
    \centering
    \includegraphics[width=.95\textwidth, trim=0cm 0cm 0cm 1.15cm, clip=TRUE]{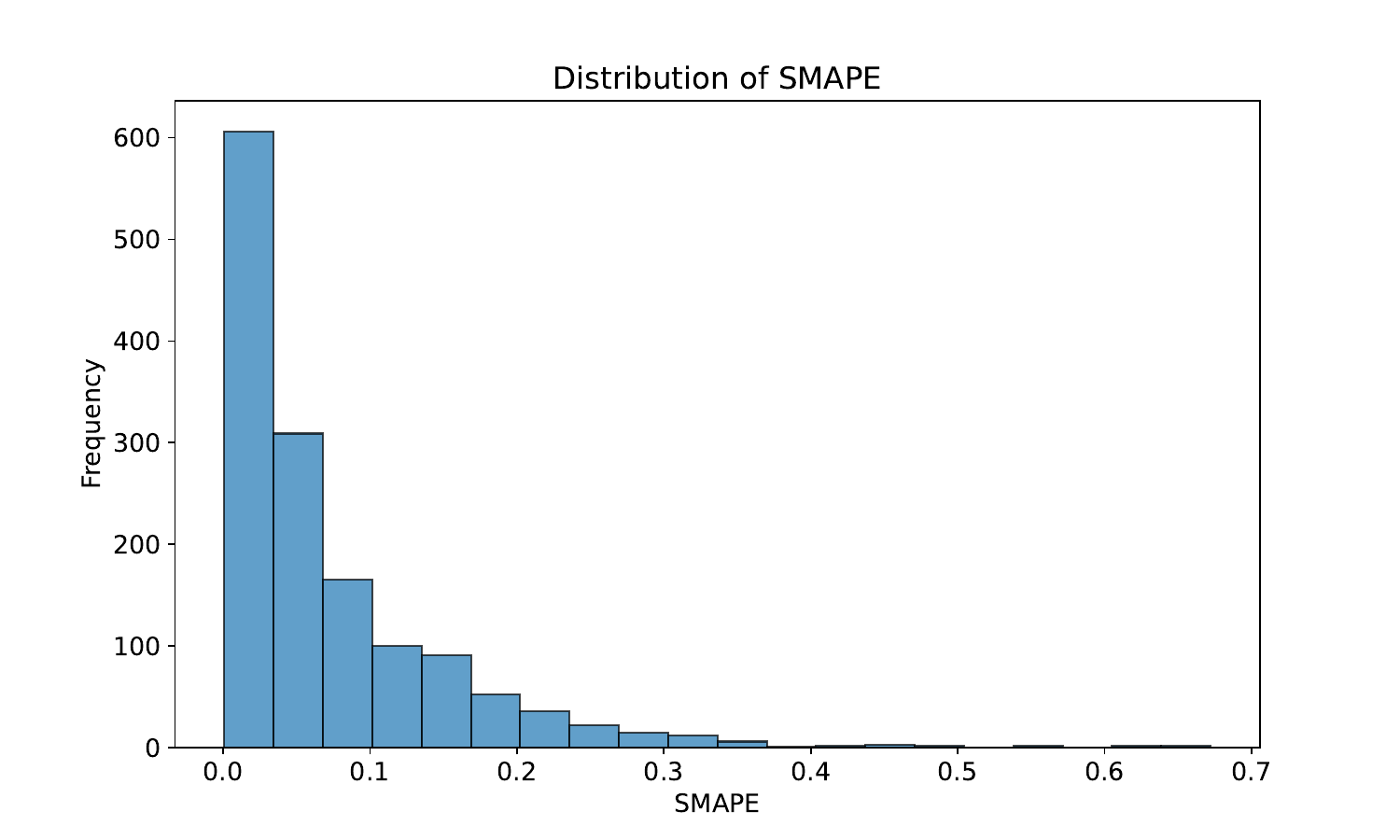}
    \caption{Forecasting performance of a model across several univariate time series according to SMAPE.}
    \label{fig:performance_distribution}
\end{figure}

Figure \ref{fig:performance_distribution} illustrates the forecasting performance of a model across several univariate time series. It shows the distribution of SMAPE (symmetric mean absolute percentage error) incurred by a forecasting model across 1428 time series. The distribution is right-skewed, where the long-tail denotes time series where the model is under stress and exhibits larger errors. 
We will provide further details on this figure in Section \ref{sec:experiments}.

The goal of this paper is to stress test a given forecasting model. Particularly, identify the conditions that lead to a large error in a given time series.
Understanding these factors is an essential step to improve the understanding of the inner workings of forecasting models. Ultimately, this knowledge promotes a more reliable and transparent use of forecasting systems by enabling end-users to adapt their decisions under the identified stress conditions.
We accomplish this goal by proposing a new approach for modelling stress in univariate time series forecasting models called \texttt{MAST} (short for Meta-learning and data Augmentation for Stress Testing). 

Meta-learning involves using machine learning to improve the performance of machine learning models, with tasks such as algorithm selection \cite{brazdil2022metalearning} or dynamic model combination \cite{cerqueira2019arbitrage}. \texttt{MAST} uses meta-learning to model the occurrence of large errors based on structural properties of time series (e.g., seasonal strength). These structural properties are computed using time series feature extraction methods \cite{montero2020fforma,barandas2020tsfel} and include statistics such as the average value or standard deviation.

Large errors (relative to the typical performance of a model) represent a small fraction of the dataset (c.f. Figure \ref{fig:performance_distribution}). A limited number of samples complicates the process of modelling these instances. In this context, we frame the above meta-learning problem as an imbalanced learning task.
Imbalanced domain learning is the field of machine learning devoted to modelling datasets where the target distribution is skewed, which means most observations belong to a single class. However, the minority class is often the most relevant, which, in our cases, represents the cases where a model incurs a large error. Resampling techniques are typically used to tackle class imbalance in binary classification problems. 
We explore resampling methods (e.g., \texttt{SMOTE} \cite{chawla2002smote}) to generate synthetic samples concerning large errors. We hypothesize that the augmented dataset improves the performance of the meta-learning model.

We validate the proposed approach in three benchmark datasets that comprise a total of 49.794 time series. The experiments suggest that our method is able to identify the conditions that lead to large errors by a forecasting model. Using Shapley values \cite{lundberg2017unified}, we provide concrete examples about the conditions where a given model performs poorly.
The proposed method and experiments are available online \footnote{\url{https://github.com/ricardoinaciopt/mast}}.

In summary, the contributions of this paper are the following:
\begin{itemize}
    \item A novel methodology for stress testing forecasting models based on meta-learning and data augmentation called \texttt{MAST};

    \item A data augmentation technique based on resampling strategies that creates synthetic observations concerning large forecasting errors;

    \item A set of extensive experiments that validate the proposed method in benchmark datasets, including usage examples based on Shapley values. 
\end{itemize}

\section{Literature Review}\label{sec:background}

This section lays out the foundations of our research. We start by defining the problem of univariate time series forecasting  (Section \ref{sec:problemdef}), focusing on machine learning approaches trained in a global manner. Then, we describe several methods for data augmentation in Section \ref{sec:rw_data_augment}. We highlight practices designed for time series and approaches tailored for tackling class imbalance.
We also list several frameworks for time series feature extraction (Section \ref{sec:featureextraction}).
Finally, we briefly overview the relevance of stress testing and its scope within responsible artificial intelligence (Section \ref{sec:responsible_ai}).

\subsection{Problem definition}\label{sec:problemdef}

\subsubsection{Univariate time series forecasting.}\label{sec:ar}

A univariate time series is a time-ordered sequence of observations. It can be defined as $Y = \{y_1, y_2, \dots,$ $y_t \}$, where $y_i \in \mathbb{R}$ is the numeric value of $Y$ collected at time $i$ and $t$ is the length of $Y$.
In this work, we aim at predicting the value of the next $h$ observations of a given univariate time series, $y_{t+1}, \ldots, y_{t+h}$, based on historical data.

We address the forecasting task using machine learning approaches and an auto-regressive methodology.
We use sliding windows based on time delay embedding to reconstruct a time series $Y$ for supervised learning. Each observation of a time series is modelled as a function of its past $p$ lags~\cite{bontempi2013machine}.
Essentially, we build a dataset $D=\{<X_i, y_i>\}^t_{i=p+1}$ where $y_i$ represents the $i$-th observation and $X_i \in \mathbb{R}^p$ is the $i$-th corresponding set of $p$ lags: $X_i = \{y_{i-1}, y_{i-2}, \dots, y_{i-p} \}$. The goal is to build a regression model $f$ with the form $y_i = f(X_i)$. 

\subsubsection{Global Forecasting Models.}

We focus on forecasting problems that involve a collection of  multiple univariate time series: $\mathcal{Y} = \{Y_1, Y_2, \dots, Y_n\}$, where $Y_j$ is the $j$-th time series and $n$ is the number of time series in the collection. 

Local forecasting methods build an individual model for each time series in a collection $\mathcal{Y}$. Classical techniques, such as ARIMA \cite{hyndman2018forecasting}, tend to follow this approach.
In contrast, recent machine learning methods applied to forecasting tend to follow a global approach and build a model for a given collection using the historical data of all available time series. Global forecasting models have outperformed local ones in benchmark datasets and competitions \cite{godahewa2021ensembles}.

Using several time series to train a forecasting model has been shown to lead to better forecasting performance \cite{godahewa2021ensembles,cerqueira2024fly}. In practice, this is achieved by transforming each time series using the approach described in Section \ref{sec:ar}. Then, a global model is trained on the concatenation of the resulting datasets $\mathcal{D}=\{D_1, D_2, \dots, D_n\}$, where $D_j$ is the dataset built for time series $Y_j$. 
A global forecasting approach often requires appropriate data preprocessing steps such as transformations, which we detail in Section \ref{sec:experiments}.

\subsection{Data Augmentation}\label{sec:rw_data_augment}

Data augmentation techniques are useful when a large or diverse enough dataset is not available for training a model \cite{bandara2021improving}. For time series, there are various methods to create synthetic time series. These range from simple methods such as jittering or flipping the original time series \cite{wen2020time}, to more sophisticated approaches such as seasonal decomposition \cite{cleveland1990stl} or generative models \cite{kang2020gratis}. 

Most time series data augmentation techniques aim to create synthetic time series that preserve the properties of the original ones. There are also approaches that attempt to create time series samples in parts of the data space with limited examples. For instance, Cerqueira et al. \cite{cerqueira2024time} leverage resampling techniques such as \texttt{SMOTE} \cite{chawla2002smote} to increase the sample size of particular time series within a collection $\mathcal{Y}$. \texttt{SMOTE} randomly selects instances from the minority class (in the case of Cerqueira et al. \cite{cerqueira2024time}, individual time series in a collection), and interpolates them based on their nearest neighbours to create new samples. 
While resampling methods are typically used to mitigate class imbalance issues in classification, Cerqueira et al. \cite{cerqueira2024time} apply them in forecasting problems to improve the global-local trade-off \cite{januschowski2020criteria}.

Overall, the ultimate goal of existing time series data augmentation techniques is to obtain an enriched dataset that leads to better forecasting performance \cite{cerqueira2024fly}. In this work, we leverage these methods to improve meta-learning and stress testing a forecasting model. We explore oversampling strategies, to generate synthetic samples in observations where a forecasting model shows a large error, increasing the overall representation of such instances. We focus specifically on \texttt{SMOTE} and \texttt{ADASYN} as they are well-established methods in the literature of imbalanced domain learning \cite{moniz2021automated}.

\subsection{Time Series Feature Extraction}\label{sec:featureextraction}

Extracting features from time series has been shown to improve performance in different tasks, such as forecasting \cite{lemke2010meta} or classification \cite{barandas2020tsfel}. Several frameworks have been proposed for this task. Examples include the works by Prudêncio and Ludermir~\cite{prudencio2004using}, Lemke et al.~\cite{lemke2010meta}, Montero-Manso et al. \cite{montero2020fforma}, Barandas et al.~\cite{barandas2020tsfel}, or Christ et al.~\cite{christ2016distributed}.

In the context of time series forecasting, feature-based modelling is typically used in meta-learning approaches for model selection and combination. This process involves summarising each time series in a collection into a set of features. Then, a metamodel is trained to predict the most appropriate learning algorithm based on these features. For a new query time series, this metamodel predicts which model, or set of models, should be used for forecasting. In the benchmark M4 forecasting competition, Montero-Manso et al. \cite{montero2020fforma} applied this approach and ranked second place, outperforming several classical and machine learning approaches. 
Besides forecasting, time series feature extraction is a common approach for time series classification tasks. In these, the label associated with a given time series is modelled based on a set of features \cite{barandas2020tsfel}.

In this work, we use the \textit{tsfeatures} framework for time series feature extraction \cite{montero2020fforma}. As mentioned above,  Montero-Manso et al. \cite{montero2020fforma} developed a competitive feature-based forecasting approach using this framework. Overall, \textit{tsfeatures} contains a set of 42 features, including entropy, linearity, lumpiness, or trend. We refer the reader to the work by Montero-Manso et al. \cite{montero2020fforma} for a comprehensive description of these.

\subsection{Stress Testing}\label{sec:responsible_ai}

Building reliable and robust predictive models is a key step towards the responsible use of machine learning \cite{freiesleben2023beyond}. This task is challenging because real-world data is susceptible to \textit{stress} conditions that undermine the performance of models. These conditions can be caused by issues such as out-of-distribution data, missing values, or an inadequate inductive bias concerning some parts of the input space.

Identifying the conditions in which a model is under stress and is prone to under-perform is a crucial aspect for supporting the adoption of machine learning systems, especially in sensitive domains such as healthcare. Stress testing is an emerging topic that addresses this problem \cite{10.1007/978-3-031-30047-9_8}. The goal of stress testing is to evaluate the reliability of machine learning approaches by identifying, modelling and simulating challenging scenarios. 

Stress testing was recently pioneered in the context of computer vision in the work by Cunha et al. \cite{10.1007/978-3-031-30047-9_8}. They propose an extension of generative adversarial networks called GASTeN (Generative Adversarial Stress Testing Network) that aims to create synthetic samples with two key properties: i) realistic and ii) challenging, where a given classifier serves a prediction with low confidence. Their framework enables the simulation of a diverse set of conditions where a model is not reliable, which provides useful insights for end-users. While they focus on generating samples that are challenging for a given model, we adopt a meta-learning approach to model and identify the weaknesses of a model. To our knowledge, this is the first work that explores meta-learning for stress testing. 

Similar to us, Rožanec et al. \cite{app11199243} also address the problem of identifying poor forecasts made by a global model, leveraging anomaly detection methods to this effect. However, they focus on managing the global-local trade-off and improving forecasting performance. 

\section{Stress Testing Forecasting Models using Meta-learning}

\begin{figure}
    \centering
    \includegraphics[width=0.8\linewidth]{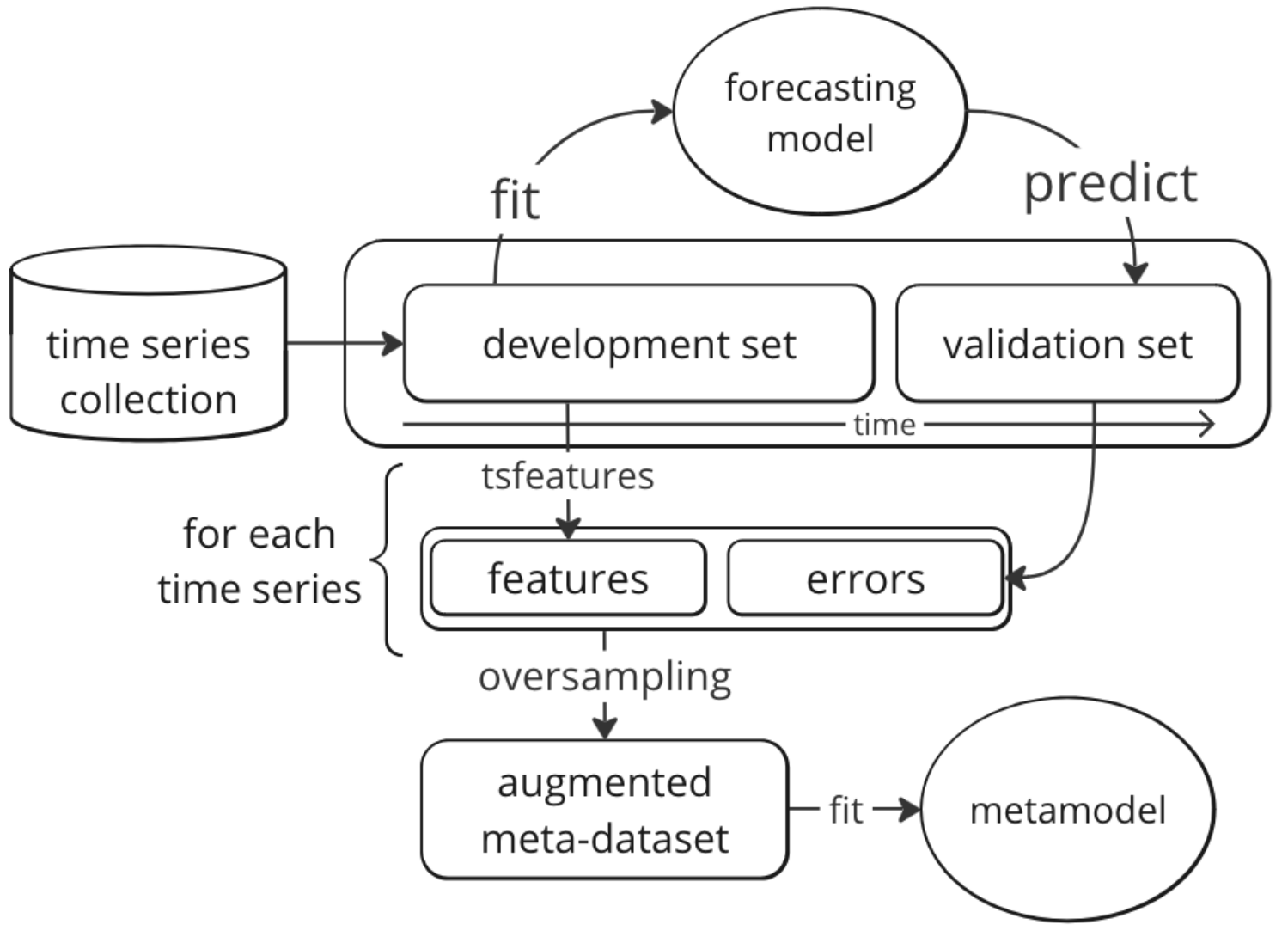}
    \caption{Workflow behind \texttt{MAST}, which is split into a development stage and an inference stage. In the development stage, we conduct performance estimation, feature extraction, data augmentation, and meta-learning. Then, the resulting metamodel is applied during the inference stage to predict whether a forecasting model will incur a large error.}
    \label{fig:dastfm-diagram}
\end{figure}

This section formalizes \texttt{MAST}, the proposed approach for modelling the stress based on meta-learning. 
We address univariate time series forecasting problems using datasets involving multiple time series. We tackle this task using an autoregressive approach, as defined in Section \ref{sec:problemdef}.
In this work, \texttt{MAST} involves using meta-learning for modelling the performance of a forecasting model based on time series features and data augmentation.
The workflow of \texttt{MAST}, which is illustrated in Figure \ref{fig:dastfm-diagram}, is split into two stages: i) a development stage and; ii) an inference stage.

\subsection{Development stage}
In the development stage, we aim at building a metamodel that predicts the probability that a time series will induce a large error on a forecasting model.
This goal is split into four steps:
\begin{enumerate}
    \item Performance estimation: Conducting a training plus evaluation cycle to estimate the performance of a forecasting model using a collection of time series;

    \item Feature extraction: Using the training data from step 1., summarize each time series into a set of features;

    \item Data augmentation: Create synthetic time series samples to enrich the meta-dataset concerning large errors;

    \item Meta-learning: Build a metamodel that predicts the performance of a forecasting model based on time series features.
\end{enumerate}

\subsubsection{Performance estimation.}

We conduct a procedure to estimate the performance of a forecasting model on each time series in the available collection $\mathcal{Y}$. 
Let $\mathcal{D}$ denote the dataset built from $\mathcal{Y}$, which is prepared for auto-regression. We split $\mathcal{D}$ into training and testing sets, $\mathcal{D}_{\text{train}}$ and $\mathcal{D}_{\text{test}}$, and fit a model $f$ using $\mathcal{D}_{\text{train}}$. The performance of $f$ is evaluated on $\mathcal{D}_{\text{test}}$, resulting in error scores according to SMAPE metric. This leads to a set of performance estimates $E = \{e_1, e_2, \dots, e_n \}$, where $e_i$ is the performance estimate of $f$ for time series $Y_i$.

\subsubsection{Feature extraction and meta-dataset.}

We summarise the training set of each time series $Y \in \mathcal{Y}$ using feature extraction. 
Let $g$ denote a feature extraction function that transforms a time series $Y_i$ into a set of features $Z_i = \{z_{i,1}, z_{i,2}, \dots, z_{i,m}$\}, where $m$ is the number of features and $z_{i,k}$ is the $k$-th feature of $Y_i$.
This transformation is written as $Z_i = g(Y_i)$.
There are several approaches in the literature for the function $g$, including TSFEL \cite{barandas2020tsfel}, tsfeatures \cite{montero2020fforma}, or catch22 \cite{lubba2019catch22}. These involve several statistical operations, such as the mean or standard deviation of the input time series.
While the methodology is applicable to any of these approaches, we focus on \textit{tsfeatures} \cite{montero2020fforma}, as we mentioned in Section \ref{sec:featureextraction}.

The results from the performance estimation, $E$, and feature extraction $Z$ are concatenated into a meta-dataset $\mathcal{D}' = \{<Z_i, e_i >\}^{n}_{i=1}$.
Each element in $\mathcal{D}'$ includes the features of a given time series, along with the error score of a given forecasting model $f$ on that data. Note that the features are extracted using the training data, while the error is computed on the test data. This process preserves the temporal order of data and avoids data leakage. 

Our goal is to use the meta-dataset to develop a metamodel $f'$ that classifies where the forecasting model $f$ will incur a large error in a given time series. 
We define a binary variable for large errors based on the error scores $E$, which is determined as follows:
\begin{equation}\label{eq:eq}
  b_i = \begin{cases}
            1 & \text{if } e_i > \tau ,\\
            0 & \text{otherwise}.
          \end{cases}
\end{equation}

\noindent For a given time series, $b_i$ takes the value of 1 if $e_i$ is above $\tau$, which represents the threshold above which an error is considered large. The threshold $\tau$ can be defined by domain expertise or in a data-driven way, e.g. using percentiles.
In effect, the meta-dataset $\mathcal{D}' = \{<Z_i, b_i >\}^{n}_{i=1}$ involves a binary classification task. The goal is to model upcoming large errors ($b$) based on time series features ($Z$) extracted from historical data.

\subsubsection{Data augmentation.}

We work under the assumption that large errors are rare. Thus, the distribution of $b$ is imbalanced, with $b=1$ being the minority class.
The meta-class imbalance poses an additional challenge for learning an adequate model for large errors. We tackle this issue by framing the task as an imbalanced domain learning problem \cite{moniz2021automated}, and use resampling approaches for data augmentation.

We employ an oversampling algorithm (e.g. \texttt{SMOTE} \cite{chawla2002smote}) to create new \newline ($Z_{i, synthetic}$, $b_{i, synthetic}$) samples concerning large errors ($b=1$).
The idea is to simulate conditions that lead to large errors based on the existing data.

\subsubsection{Meta-learning}

The created synthetic samples are represented in a dataset $\mathcal{D}'_{synthetic}$, which is concatenated with the original dataset $\mathcal{D}'$ to augment it. Subsequently, a metamodel $f'$ is trained based on the combined meta-dataset $\mathcal{D}' \cup \mathcal{D}'_{synthetic}$. 
The output of the metamodel $\hat{b}_i = f'(Z_i)$ provides information about the probability that the forecasting model $f$ will incur a large error on $Y_i$.

\subsection{Inference stage}\label{sec:inferencestage}

In the inference stage, we use the metamodel $f'$ to predict the probability that $f$ will incur a large error for a query time series $Y_{query}$. We conduct this process as follows:
\begin{enumerate}
    \item Feature extraction: Transform $Y_{query}$ into a set of features $Z_{query}$

    \item Compute the probability of a large error using the metamodel: $\hat{b} = f'(Z_{query})$
\end{enumerate}

\noindent The information about the probability of a large error can be used in various ways, for example deferring the forecasts by not accepting the predictions made by $f$. Another possibility is to use the metamodel output for a descriptive analysis by characterizing the feature space in conditions with high probability of a large error.

\section{Experiments}\label{sec:experiments}

We carried out extensive experiments to evaluate the proposed method. The experiments focus on the following research questions:
\begin{itemize}
    \item \textbf{RQ1:} How does the base forecasting model compare with benchmark approaches?

    \item \textbf{RQ2:} Is the metamodel able to predict large errors by a given forecasting model?

    \item \textbf{RQ3:} What is the impact of data augmentation using resampling in the performance of the metamodel?

    \item \textbf{RQ4:} How does the performance of the metamodel vary with different values of $\tau$?
    
\end{itemize}

\noindent Besides these, we also present an example of how the results of the metamodel can be used to understand the conditions where a forecasting model is expected to perform poorly.

\subsection{Data}
\label{sec:data}

We use the following datasets in the experiments:  M3 Monthly \cite{MAKRIDAKIS2000451}, M4 Monthly \cite{makridakis2020m4}, and Tourism Monthly \cite{ATHANASOPOULOS2011822}. These datasets were part of different forecasting competitions and represent long-standing univariate time series forecasting benchmarks. In the interest of conciseness, we focus on time series with a monthly sampling frequency.  Table \ref{tab:data} provides a brief summary of the data. 

\begin{table}[htbp]
\centering
\caption{Description of the datasets, including number of time series, number of observations, and average value length.}
\begin{tabular}{lr@{\hskip 0.5cm}r@{\hskip 0.5cm}r@{\hskip 0.5cm}}
\hline
 & \# Time series & \# Observations & Avg. length \\
\hline
M3 Monthly & 1428 & 167562 & 117 \\
M4 Monthly & 48000 & 11246411 & 234 \\
Tourism Monthly & 366 & 109280 & 299 \\
\hline
\end{tabular}
\label{tab:data}
\end{table}

\noindent Both the forecasting horizon ($h$) and the number of lags ($p$) are set to 12, which represents one seasonal cycle (one year) of data. 

\subsection{Experimental Design}

\subsubsection{Performance estimation.}

\begin{figure}
    \centering
    \includegraphics[width=0.6\linewidth]{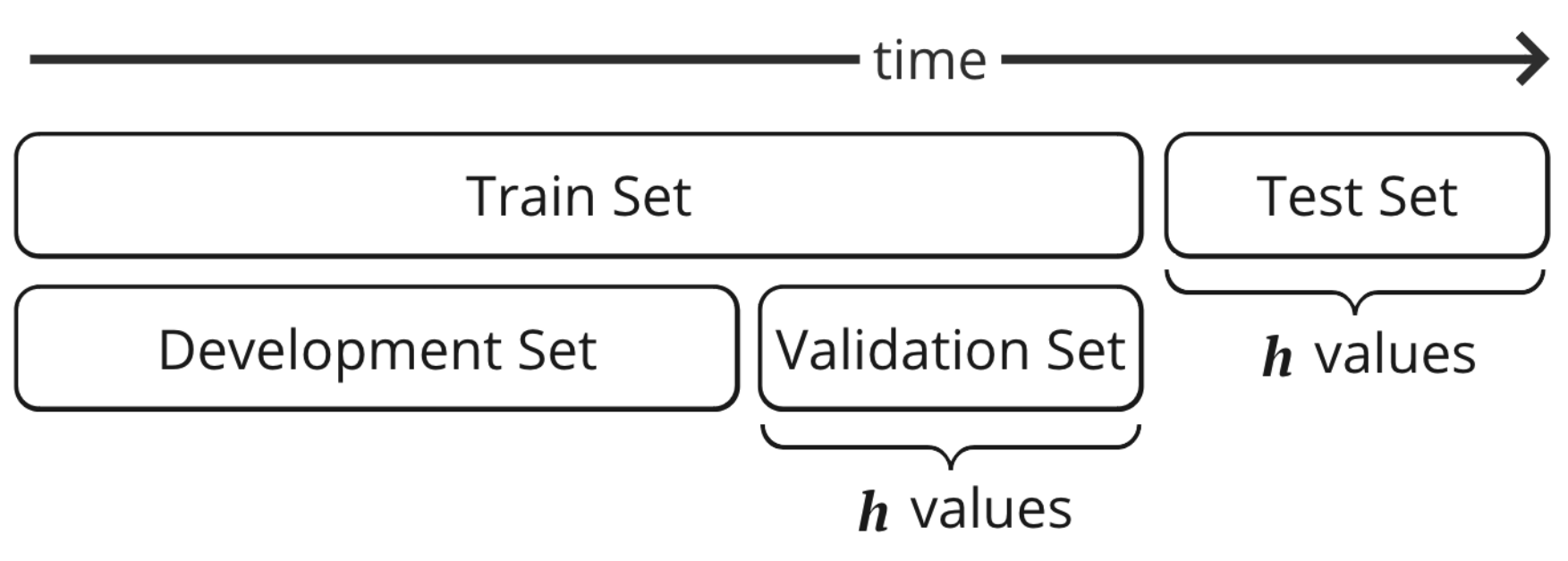}
    \caption{Data preparation step, where for each time series, a split equal to the forecasting horizon (\textit{h}) is done at the end of the data, to define the test set, assigning the rest to the train set. Another split with the same size is done at the end of the latter, defining the validation set, assigning the remaining to the development set.}
    \label{fig:experimental-design}
\end{figure}

The data partitioning process is illustrated in Figure~\ref{fig:experimental-design}.
We split the available data into training and testing sets for performance estimation. The test set is composed of the last $h$ observations of each time series in $\mathcal{Y}$.
The training set is further split into a development set and a validation set, where the latter is composed of the last $h$ observations of each time series before the test set.

We carry out the methodology for building the metamodel using the development set and the validation set (c.f. Figure \ref{fig:dastfm-diagram}).
Using the training set, we re-train the forecasting model and carry out the inference stage described in Section \ref{sec:inferencestage}. Finally, the performance of the forecasting model and the metamodel are evaluated using the test set. For each time series, we evaluate the forecasting performance of $f$. Using those results, we evaluate the ability of the metamodel $f'$ to predict large errors.
We evaluate forecasting performance using SMAPE, a benchmark metric \cite{makridakis2020m4}. Concerning the metamodel, we resort to AUC (area under the ROC curve) to evaluate its performance.

\subsubsection{Forecasting model.}\label{sec:forecastmodel}

The forecasting model is built using a \texttt{lightgbm} (\texttt{LGBM}) regression algorithm. This approach has been shown to provide state-of-the-art forecasting performance. For instance, it is the algorithm used in the winning solution of the M5 forecasting competition \cite{makridakis2022m5}. Following Bandara et al. \cite{bandara2021improving}, we train a global forecasting model for each dataset listed in Table \ref{sec:data} with the approach described in Section \ref{sec:problemdef}. 
We resorted to the \textit{mlforecast} implementation of \texttt{lightgbm} called \textit{AutoLightGBM}. This methods automatically conducts hyperparameter tuning using the \texttt{optuna} framework \cite{optuna_2019}, which uses Bayesian optimization. Specifically, the following parameters are optimized: \textit{learning rate,} \textit{number of leaves}, \textit{max depth}, and \textit{number of estimators}. 
We also include the seasonal naive method in the experiments as a benchmark. This approach forecasts the upcoming values of a time series using the last known observation of the same period.

The metamodel is also trained using a \texttt{lightgbm} algorithm. We use an objective function tailored for binary classification and optimize the hyperparameters via grid search. Moreover, we use the area under the ROC curve (AUC) as the evaluation metric for the metamodel. While AutoLightGBM provides an off-the-shelf approach for building optimized models based on \texttt{lightgbm}, it is only available for forecasting. Thus, for the binary metamodel classifier, we optimize \texttt{lightgbm} using grid search.

\subsubsection{Parameter setting.}

We define large errors in a data-driven way using percentiles. The threshold $\tau$ is set to the percentile 90 of the SMAPE performance estimates obtained in the validation set. We work under the assumption that the forecasting model typically performs arbitrarily well, except for rare instances. Nonetheless, we carry out a sensitivity analysis to analyse the effect of the threshold on the performance of the metamodel.

As mentioned before, we leverage the \textit{tsfeatures} framework for time series feature extraction \cite{montero2020fforma}. 
Concerning data augmentation, we test two popular oversampling algorithms: \texttt{SMOTE} \cite{chawla2002smote}, and \texttt{ADASYN} \cite{he2008adasyn}.

\subsection{Results}

Figure \ref{fig:performance_distribution} shows the distribution of SMAPE scores by the forecasting model described in Section \ref{sec:forecastmodel} across all the time series in the M3 dataset. 
We start by validating the performance of the forecasting model by comparing it with a seasonal naive baseline. While we aim at identifying the weaknesses of a forecasting model, ensuring that it outperforms a baseline is important for its applicability. 
The results are reported in Table \ref{tab:results1}, which show the SMAPE scores of each approach in each dataset. The developed model outperforms seasonal naive in all cases, which validates its forecasting performance (\textbf{RQ1}).

\begin{table}
\caption{Forecasting performance of each method using SMAPE. Bold font denotes the best approach.}
\centering
\label{tab:results1}
\begin{tabular}{lr@{\hskip 0.5cm}r}
\toprule
 & \texttt{Seasonal Naive} & \texttt{LGBM} \\
\midrule
M3 Monthly & 7.98\% & \textbf{7.28\%} \\
M4 Monthly & 7.17\% & \textbf{6.58\%} \\
Tourism Monthly & 9.15\% & \textbf{8.50\%} \\
\bottomrule
\end{tabular}
\end{table}

\begin{table}
\caption{metamodel performance in each dataset according to AUC. Bold font denotes the best method in each dataset.}
\centering
\label{tab:results2}
\begin{tabular}{lr@{\hskip 0.5cm}r@{\hskip 0.5cm}r@{\hskip 0.5cm}}
\toprule
& \texttt{ADASYN} & \texttt{SMOTE} & \texttt{No Augmentation}\\
\midrule
M3 Monthly & \textbf{0.728} & 0.678 & 0.516 \\
M4 Monthly & \textbf{0.713} & 0.709 & 0.566 \\
Tourism Monthly & \textbf{0.787} & 0.761 & 0.730 \\
\bottomrule
\end{tabular}
\end{table}

The results concerning the metamodel are reported in Table \ref{tab:results2}.
The metamodel is able to predict large errors with a reasonable performance (\textbf{RQ2}). Overall, it performs better when coupled with data augmentation using \texttt{ADASYN}. When no data augmentation is applied, the proposed method struggles in two of the datasets (M3 and M4), which emphasizes the importance of the synthetic samples (\textbf{RQ3}).

\begin{figure}
    \centering
    \includegraphics[width=\linewidth]{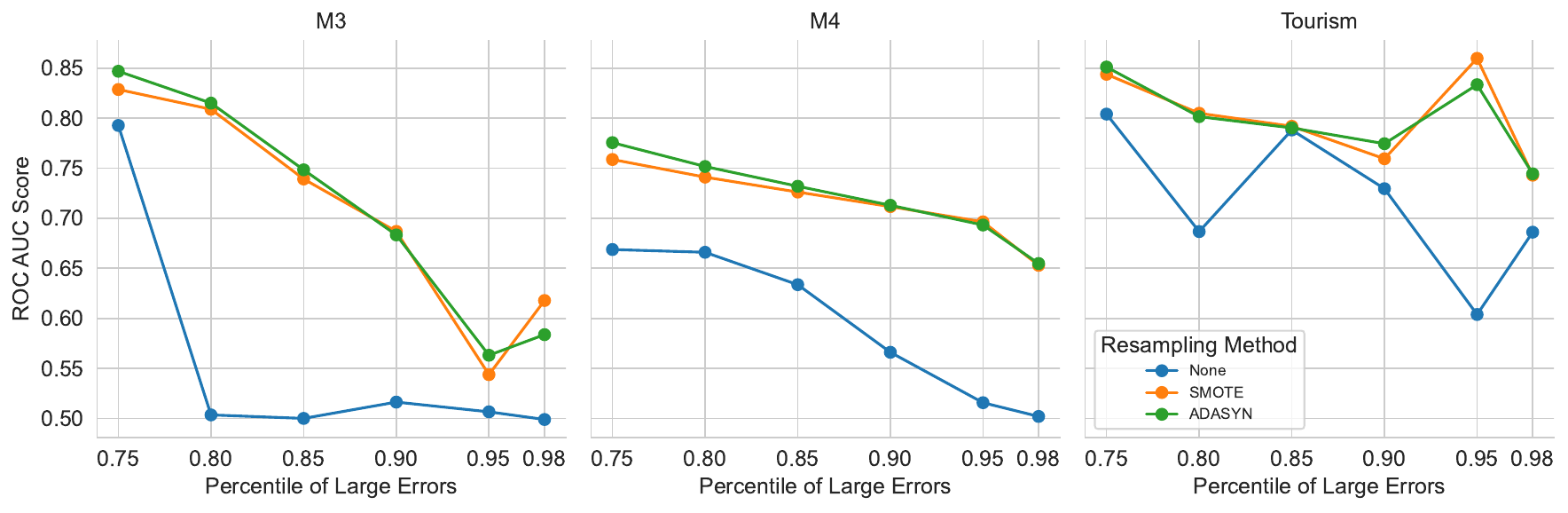}
    \caption{Analysis of the impact of error percentile threshold in metamodel performance, measured in AUC. For each time series collection, the performance for the base and augmented metamodels is compared regarding six levels of percentile error threshold.}
    \label{fig:roc-percentiles}
\end{figure}

We conducted a sensitivity analysis to assess the impact of the threshold for large errors ($\tau$) in the results (\textbf{RQ4}). The results are shown in Figure \ref{fig:roc-percentiles}, which illustrates the AUC for increasing values of $\tau$ for each dataset and approach. Overall, increasing the value of $\tau$ leads to worse performance. In all three datasets, the metamodel trained on augmented metadata shows better performance than a variant without data augmentation.

\subsection{Metamodel explanations}

The results presented in the previous section provide evidence that \texttt{MAST} is able to detect large errors. We can leverage Shapley values \cite{lundberg2017unified} to explore how the time series features affect the output of the metamodel.
For conciseness, this analysis is focused on the M3 dataset and on the five most important features on this task, according to the metamodel. These are the following (most important first): \texttt{linearity}, \texttt{lumpiness}, \texttt{hw\_beta}, \texttt{spike}, and \texttt{hw\_gamma}. Again, we refer to the work by Montero-Manso et al. \cite{montero2020fforma} for a description of these\footnote{Also available here: \url{https://htmlpreview.github.io/?https://github.com/robjhyndman/M4metalearning/blob/master/docs/M4_methodology.html#features}}.

\begin{figure}
    \centering
    \includegraphics[width=0.9\linewidth]{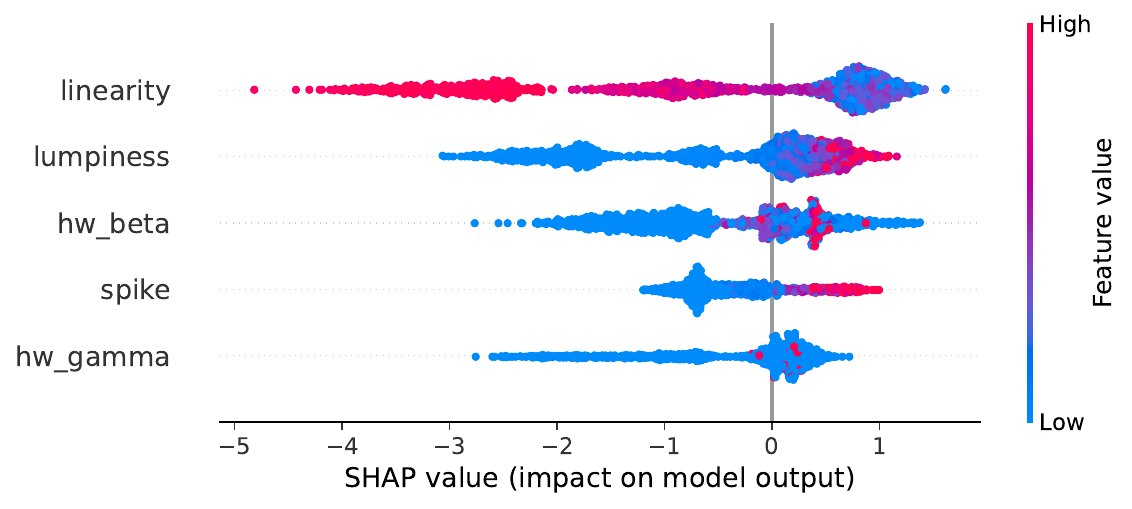}
    \caption{Shapley values for the 5 most relevant features for the M3 dataset's metamodel, indicating which ones contribute positively and negatively to its output. Values to the right of the vertical line indicate a positive influence, while the ones at the left indicate a negative one. The colour indicates the magnitude of the features' value.}
    \label{fig:features-shap}
\end{figure}

Figure \ref{fig:features-shap} shows a summary plot of the Shapley values for the 5 most relevant features for the M3 dataset.
This plot shows interesting insights about the performance of the forecasting model based on the predictions of the metamodel. For example, high \texttt{linearity} values reduces the probability of classifying a time series as large error inducing. This suggests that the forecasting model may struggle in time series with a low \texttt{linearity} value. 

\section{Discussion}

This work addresses the problem of evaluating the robustness of univariate time series forecasting models to time series with varying characteristics.
We use meta-learning and data augmentation to model the relationship between large errors and the structural properties of time series. 
The proposed approach was validated using benchmark datasets composed of monthly time series. The results indicate that our approach is able to identify conditions that lead to large errors. The conditions identified by the metamodel can be explored using techniques such as Shapley values.

Overall, the experiments suggest that using meta-learning to model large errors is a promising approach for improving the transparency and reliability of forecasting models.
We plan to extend the proposed method in future work, including the following aspects:
\begin{itemize}
    \item modelling stress in the form of high uncertainty, as opposed (or a complement) to stress in terms of large errors;

    \item Model the stress of algorithms instead of trained models to understand which methods should be used for a given time series dataset;

    \item Test different data augmentation approaches, such as moving blocks bootstrapping \cite{bandara2021improving}.
    
    \item Apply the developed approach in a wider range of time series, including different sampling frequencies.
\end{itemize}

\section{Final Remarks}

modelling the stress of forecasting models, such as large errors, is important to improve their transparency and reliability. This paper introduces a novel framework for modelling and describing the stress in univariate time series forecasting models based on meta-learning and data augmentation.

We validated our approach using benchmark datasets that contain multiple univariate time series. The developed metamodel is able to relate large errors with structural properties of time series, such as \textit{linearity} or \textit{spike}. In future work, we plan to extend our approach for different time series and stress scenarios.
The code and data is publicly available\footnote{c.f. footnote 4}.

\bibliographystyle{splncs04}
\bibliography{mybibliography}

\end{document}